\documentclass[a4paper,11pt,article,oneside,table,xcdraw]{memoir}
\usepackage{ecis2019}
\usepackage[acronym]{glossaries}
\usepackage{diagbox}
\usepackage{hhline}
\usepackage{pdflscape}
\usepackage{eurosym}
\usepackage{graphicx}
\usepackage{booktabs}
\usepackage{todonotes}
\usepackage{dsfont}
\makeatletter
\presetkeys%
    {todonotes}%
    {inline}{}%
\makeatother
\usepackage{hyperref}
\usepackage{breakcites}
\usepackage{array}
\usepackage{amsmath}
\usepackage{multirow}
\usepackage{rotating}
\usepackage{makecell}
\usepackage[acronym]{glossaries}
 \usepackage[normalem]{ulem}
 \useunder{\uline}{\ul}{}
\usepackage{amssymb}

\newacronym{pbpm}{PBPM}{predictive business process monitoring}
\newacronym{bpm}{BPM}{business process management}
\newacronym{dsr}{DSR}{design science research}
\newacronym{dnn}{DNN}{deep neural network}
\newacronym{lstm}{LSTM}{long short-term memory}
\newacronym{bilstm}{BiLSTM}{bi-directional long short-term memory}
\newacronym{ml}{ML}{machine learning}
\newacronym{dl}{DL}{deep learning}
\newacronym{pm}{PM}{process mining}
\newacronym{it}{IT}{information technology}
\newacronym{is}{IS}{information systems}
\newacronym{pdp}{PDP}{partial dependence plots}
\newacronym{ebm}{EBM}{explainable boosting machine}
\newacronym{gbm}{GBM}{gradient boosting machine}

\newacronym{xai}{XAI}{explainable artificial intelligence}
\newacronym{shap}{SHAP}{Shapley additive explanations}
\newacronym{lime}{LIME}{local interpretable model-agnostic explanations}
\newacronym{gam}{GAM}{generalized additive model}
\newacronym{gaim}{GAIM}{generalized additive index model}
\newacronym{nam}{NAM}{neural additive model}
\newacronym{xnn}{xNN}{explainable neural network}
\newacronym{exnn}{ExNN}{enhanced explainable neural network}
\newacronym{ppr}{PPR}{projection pursuit regression}
\newacronym{gaminet}{GAMI-Net}{GAMI-Net}
\newacronym{igann}{I-GANN}{interpretable generalized additive neural network}
\newacronym{vi}{VI}{variable importance}
\newacronym{ice}{ICE}{individual conditional expectation}
\newacronym{rf}{RF}{random forest}
\newacronym{dt}{DT}{decision tree}
\newacronym{svm}{SVM}{support vector machine}
\newacronym{lr}{LR}{logistic/linear regression}
\newacronym{xgb}{XGB}{extreme gradient boosting}
\newacronym{mlp}{MLP}{multi-layer perceptron}

\newacronym{rmse}{RMSE}{root mean square error}
\newacronym{mae}{MAE}{mean absolute error}
\usepackage{enumitem}
\setlist[description]{leftmargin=.25cm,labelindent=\parindent}
\newcolumntype{L}[1]{>{\raggedright\let\newline\\\arraybackslash\hspace{0pt}}m{#1}}
\newcolumntype{C}[1]{>{\centering\let\newline\\\arraybackslash\hspace{0pt}}m{#1}}
\newcolumntype{R}[1]{>{\raggedleft\let\newline\\\arraybackslash\hspace{0pt}}m{#1}}

\maintitle{GAM(e) Changer or Not? An Evaluation of Interpretable Machine Learning Models based on Additive Model Constraints} 

\shorttitle{Zschech et al. / Comparative Evaluation of Interpretable \gls{ml} Models} 
\category{Research Paper} 

\authors{
  Patrick Zschech, FAU Erlangen-N\"urnberg, N\"urnberg, Germany, patrick.zschech@fau.de\par
  Sven Weinzierl, FAU Erlangen-N\"urnberg, N\"urnberg, Germany, sven.weinzierl@fau.de\par
  Nico Hambauer, FAU Erlangen-N\"urnberg, N\"urnberg, Germany, nico.hambauer@fau.de\par
  Sandra Zilker, FAU Erlangen-N\"urnberg, N\"urnberg, Germany,
  sandra.zilker@fau.de\par
  Mathias Kraus, FAU Erlangen-N\"urnberg, N\"urnberg, Germany, mathias.kraus@fau.de}

\shortauthors{} 

\begin{document}
\begin{abstract}\noindent
The number of information systems (IS) studies dealing with explainable artificial intelligence (XAI) is currently exploding as the field demands more transparency about the internal decision logic of machine learning (ML) models. However, most techniques subsumed under XAI provide post-hoc-analytical explanations, which have to be considered with caution as they only use approximations of the underlying ML model. Therefore, our paper investigates a series of intrinsically interpretable ML models and discusses their suitability for the IS community. More specifically, our focus is on advanced extensions of generalized additive models (GAM) in which predictors are modeled independently in a non-linear way to generate shape functions that can capture arbitrary patterns but remain fully interpretable. In our study, we evaluate the prediction qualities of five GAMs as compared to six traditional ML models and assess their visual outputs for model interpretability. On this basis, we investigate their merits and limitations and derive design implications for further improvements.
\end{abstract}


\setlength{\parskip}{.25em}


\chapter{Introduction}
\label{chap:intro}

Due to the technological achievements in the field of \gls{ml}, many tasks related to predictive decision-making are increasingly supported by \gls{ml} models \citep{kraus_deep_2020, janiesch_machine_2021}. Prominent examples can be found in business process monitoring \citep{heinrich_process_2021, stierle_technique_2021}, hate speech detection \citep{zinovyeva_antisocial_2020}, medical diagnosis \citep{mckinney_international_2020}, industrial maintenance \citep{kraus_forecasting_2019, zschech_prognostic_2019}, or natural disaster detection \citep{devries_deep_2018}. However, most advanced \gls{ml} models such as \glspl{dnn} and \glspl{gbm} represent complex mappings between input features and the prediction targets. This results in black-box behavior which does not allow for sufficient model analysis to assess the internal decision logic and consequently leads to a lack of trust \citep{miller_explanation_2019, thiebes_trustworthy_2021}. For this reason, there is increasing concern about using such models in critical decision-making scenarios, such as healthcare, finance, and criminal justice \citep{rudin2019stop, agarwal_interpretable_2020}.

To overcome such challenges, various techniques have been proposed in recent years to make black-box behavior more transparent and comprehensible. A commonly applied solution is to provide additional model explainability by approaches like \gls{lime} \citep{ribeiro_why_2016} and \gls{shap} \citep{lundberg_local_2020}. Such techniques offer post-hoc explanations to better understand the predictions made by complex models \citep{barredo_arrieta_explainable_2020}. Nevertheless, post-hoc explanations have to be considered with caution as they can only be used to explain predictions of instances which are already known and therefore try to reconstruct the cause of a generated prediction through approximation. Hence, they might not be reliable and can result in misleading conclusions \citep{rudin2019stop, babic_beware_2021, john-mathews_critical_2021}.

Another way of providing transparency is to build intrinsically interpretable \gls{ml} models \citep{du_techniques_2019}.  In general, the structure of interpretable models is in some way constrained, such that the resulting models provide a better understanding of how predictions are generated. This may include various constraints such as linearity, additivity, smoothness, and other types of structural simplifications \citep{rudin2019stop, sudjianto_unwrapping_2020}. Widely known representatives of this class are linear/logistic regression models or decision rules, which can be easily interpreted but are too restrictive to capture more complex effects.

Beyond that, however, there are also more powerful models that are able to better address the trade-off between model accuracy and model transparency. One of the most promising directions in this area is that of \textit{\glspl{gam}} \citep{hastie_generalized_1986, lou_intelligible_2012}. In \glspl{gam}, input variables are mapped independently of each other in a non-linear manner and the mappings are summed up afterward \citep{barredo_arrieta_explainable_2020}. Thus, \glspl{gam} include additive constraints yet drop the linearity constraint. In this way, it is still possible to extract univariate mappings from the model, which show the relation between single input features and the response variable. These relations are commonly known as \textit{ridge} or \textit{shape functions} \citep{lou_intelligible_2012}. Since shape functions can take any arbitrary form, \glspl{gam} often achieve much better prediction accuracies as compared to simple linear models. However, as they do not contain high-order interactions between features, they can be easily interpreted by model users and developers for decision assessment or model debugging purposes \citep{lou_intelligible_2012}.

Traditionally, shape functions in \glspl{gam} have been learned via splines \citep{hastie_generalized_1986}, whereas recent proposals consider more advanced techniques, such as bagged and boosted tree ensembles \citep{lou_intelligible_2012, caruana2015intelligible} or specific neural networks \citep{agarwal2021neural}. These approaches allow for a higher degree of flexibility and therefore are able to achieve better predictive performance while remaining fully interpretable. Beyond that, more advanced approaches have been proposed most recently with extensions towards \glspl{gaim} \citep[e.g.,][]{vaughan_explainable_2018, yang_enhancing_2021} and other enhancements to further improve prediction qualities and/or interpretability. Taken together, these approaches show a very promising direction towards intrinsically understandable white-box \gls{ml} models that provide a technically equivalent, but ethically more acceptable alternative to black-box models, which is of pivotal interest to the \gls{is} community.

While each \gls{gam} extension has already been evaluated in its own setting by the corresponding authors/developers, there is currently no neutral, independent, and, in particular, overarching cross-model comparison. As such, it currently lacks a comprehensive study that (i) examines the advantages and disadvantages of the individual models, (ii) reflects on the suitability for IS research and practice --- especially with regards to commonly applied \gls{ml} models, and (iii) derives implications for further improvements from a design perspective. Thus, we aim to answer the following research questions:

\begin{table}[h]
\centering
\begin{tabular}{lp{14cm}}
\textit{\textbf{RQ1:}} & \textit{Can interpretable models based on additive model constraints provide competitive prediction results as compared to traditional white-box and black-box \gls{ml} models?} \\
\textit{\textbf{RQ2:}} & \textit{How do the outputs of interpretable \gls{ml} models based on additive model constraints differ from each other objectively?} \end{tabular}
\end{table}

To answer these questions, we perform a comprehensive and systematic evaluation study, in which we apply five \gls{gam}/\gls{gaim}-based models as well as six commonly used ML-models on twelve datasets to assess and compare their predictive performance. Subsequently, we use selected datasets to outline benefits and limitations concerning the interpretability of the intrinsically interpretable models. As a result, we formulate four design principles which may guide future research on the development of interpretable \gls{ml} models and their integration into \gls{ml}-based \gls{is}. Following this line, the paper is organized as follows: In Section \ref{chap:background}, we provide an overview of the conceptual background and related work. Subsequently, we elaborate on the design of our evaluation study in Section \ref{chap:method}. In Section \ref{chap:results}, we provide a detailed presentation of the results related to the assessment of predictive performance and model interpretability. Finally, we discuss our findings, outline the potential role of \gls{gam}-based models for the \gls{is} community, and provide an outlook for future work in Section~\ref{chap:discussion}.


\chapter{Conceptual Background and Related Work}
\label{chap:background}

\section{Model Explainability and Model Interpretability}
There are basically two strands of research that are concerned with the transparency of \gls{ml} models. The first one refers to \textit{post-hoc model explainability}. It deals with techniques that provide post-hoc explanations for predictions that are made by complex black-box models in the first place (e.g., \glspl{dnn}). For this purpose, additional/external approaches are used to explain how results are generated by black-box models \citep{barredo_arrieta_explainable_2020}. Such approaches are often subsumed under the term \gls{xai} and commonly known examples are variable importance \citep{breiman_random_2001}, partial dependence plots \citep{friedman_greedy_2001}, individual conditional expectation plots \citep{goldstein_peeking_2015}, \gls{lime} \citep{ribeiro_why_2016}, and \gls{shap} \citep{lundberg_local_2020}.

Post-hoc-analytical techniques have led to insightful analysis into powerful black-box models, such as gradient boosted decision trees or \glspl{dnn}. However, these techniques can not represent the full, highly complex functioning of black-box models in a simple manner, but rather give snapshots of the functioning based on few instances. By aggregating these snapshots over multiple instances, the user can draw more general conclusions about the model, yet, these conclusions are only based on a limited number of samples. For a new instance, post-hoc analyzed models could still output unforeseeable, and potentially harmful results \citep{rudin_why_2019, babic_beware_2021, kaur_interpreting_2020}.

The second line of research refers to \textit{intrinsic model interpretability} \citep{du_techniques_2019}. This field is concerned with \gls{ml} models that are designed to be inherently interpretable, often due to their simplicity, such as given by generalized linear models, point systems, simple decision trees, decision rules, or naive Bayes classifiers \citep{lou_intelligible_2012, rudin_why_2019, yang_gami-net_2021}. To retain transparency, interpretable models are usually restricted in some way by introducing certain model structures and practical constraints, such as linearity, monotonicity, additivity, sparsity, smoothness, and near-orthogonality \citep{rudin_why_2019, sudjianto_unwrapping_2020}.

While model interpretability and explainability share the same goal of providing users with insights into how models work, they differ substantially in the underlying idea of how to achieve this goal. The research stream of model interpretability designs models in such a way, that the underlying mathematical function (which fully defines the model) is simple enough that users can access and analyze it directly. In contrast, model explainability generally does not constrain models, but rather adds another layer that attempts to simplify the function so that it is digestible by a user. Furthermore, it should be noted that in this paper, the terms explainability and interpretability are not aimed at subjective perceptual processes during cognitive reasoning, but at the functionality of a model to produce objectively comprehensible outputs.

Often, there is the erroneous belief that prediction accuracy must be strongly sacrificed for interpretability/transparency which is why complex black-box models seemed to be favored over interpretable approaches during the past decade of \gls{ml} research \citep{rudin_why_2019}. However, most recent proposals and studies have demonstrated the opposite as there are seemingly algorithmic approaches that are able to combine both aspects. One of these promising algorithmic approaches is that of \glspl{gam} and their subsequent advancements based on additive model constraints \citep{barredo_arrieta_explainable_2020}.

\section{Generalized Additive Models and Recent Advancements}

Let $D=\left(X, y\right)$ denote a training dataset, where $X=(x_{1},... ,x_{n}) \in \mathbb{R}^{N \times n}$ is the feature matrix comprising $N$ observations and $n$ features. Further, $y$ denotes the corresponding targets. A generalized additive model is then defined as:
\newcommand{\R}{\mathbb{R}}
\begin{equation}
    g(y)=f_1(x_1)+...+ f_n(x_n),
    \label{eq:1}
\end{equation}

where $g(\cdot)$ is called link function and $f_i(\cdot)$ is the shape function for a feature $x_i$. If the link function is the identity, Equation~\ref{eq:1} describes an additive model (e.g., a regression model) and if the link function is the logistic function, Equation~\ref{eq:1} describes a generalized additive model (e.g., a classification model) \citep{lou_intelligible_2012, hastie_generalized_1986}. In this paper, we consider both prediction tasks, i.e., regression problems where $y\in\mathbb{R}^N$ as well as binary classification problems where $y \in \{1,0\}^N$. Given a model $F$, let $F(X)$ denote the predictions of the model for our data points $X$. The goal in both tasks is to minimize the expected value of some loss function $L(y,\,F(X))$. The purpose of \glspl{gam} is to infer the shape functions whose aggregate composition approximates the predicted response variable. The overall structure is simply interpretable, as it allows model users and developers to verify the importance of each variable. In other words, it is directly observable how each feature, through its corresponding shape function, affects the predicted output. In Section \ref{chap:results}, we will provide several illustrative examples of such shapes when assessing the output of different \glspl{gam}. Due to the intrinsic interpretability of non-linear effects, traditional \glspl{gam} have been widely used already in various application scenarios, especially in fields related to risk assessment where trust had to be built with the user \citep{barredo_arrieta_explainable_2020}, such as finance \citep{berg_bankruptcy_2007}, healthcare \citep{caruana2015intelligible}, energy supply \citep{pierrot_short-term_2011}, geology \citep{tomic_modified_2014}, and environmental studies \citep{murase_application_2009}.

Originally, shape functions in \glspl{gam} were learned via regression \textit{splines} \citep{hastie_generalized_1986}. These are piecewise polynomial functions that can approximate complex shapes through curve fitting. However, more recent studies have shown that splines are often too smooth for real-world datasets and that higher predictive performance can be achieved using more flexible models. To this end, \citet{lou_intelligible_2012} proposed to consider bagged and boosted decision trees ensembles for fitting complex shape functions. The authors further enhanced their tree-based approach by considering pairwise interaction terms \citep{lou_accurate_2013}, and made it publicly available as an easy-to-use algorithm, known as \textit{\gls{ebm}} \citep{nori2019interpretml}.

More recently, further \gls{gam} modifications have been proposed based on \glspl{dnn}. \citet{agarwal2021neural} introduced a \textit{\gls{nam}} in which shape functions are learned via individual deep subnetworks with multiple hidden layers and specific neural units. More specifically, they introduced exponential units (ExU) for fitting jagged curves that often appear in real-world datasets. Another DNN-based approach was pursued by \citet{yang_gami-net_2021} proposing \textit{GAMI-Net}. To improve predictive performance, GAMI-Net is designed to capture pairwise interactions. This also required further model constraints, such as heredity and marginal clarity constraints, to retain structural interpretability and avoid mutual absorption between main effects and pairwise interactions. Another advanced \gls{dnn} approach was proposed by \citet{vaughan_explainable_2018} presenting \textit{\gls{xnn}}, which was further improved towards \textit{\gls{exnn}} \citep{yang_enhancing_2021}. Instead of using a simple \gls{gam} structure, both models are based on the structure of \glspl{gaim}. This structure generally violates the idea of univariate feature mappings due to an additional projection layer that fully connects all input features to the following sub-networks so that each feature can possibly have a partial contribution to all corresponding shape functions.

\section{Comparative Evaluation Studies}
Since all \gls{gam}/\gls{gaim}-based models introduced above have distinct characteristics with very specific model constraints, it is worthwhile to evaluate and compare their merits and limitations for different prediction tasks and datasets. A few authors and developers have already compared their \gls{gam} extensions to some competing models as well as traditional \gls{ml} baselines to demonstrate their proposed innovations. For example, \citet{lou_intelligible_2012} performed comprehensive experiments to compare spline-based models with tree-based \glspl{gam} and additionally considered \gls{lr} and \gls{rf} as lower and upper bound baselines, respectively. When introducing GAMI-Net, \citet{yang_gami-net_2021} compared it with \gls{ebm}, splines, and several other benchmark models, including \gls{lr}, \gls{rf}, \gls{xgb}, and \gls{mlp}, using a large number of datasets (20+). Likewise, \citet{agarwal2021neural} benchmarked their \gls{nam} against \gls{ebm} and a number of traditional approaches. However, in their study, the number of evaluated datasets was limited to a smaller amount.

Apart from that, there are only a few studies that have evaluated the properties of different additive models so far. \citet{chang_how_2021} investigated a series of \glspl{gam} including splines and tree-based approaches like \gls{ebm}. They investigated the models quantitatively and qualitatively using real-world and simulated data. \citet{hohman_gamut_2019} integrated \glspl{gam} into a visual analytics tool to examine how data scientists interact with shape functions. A slightly different perspective was taken by \citet{kaur_interpreting_2020}. They also examined how data science professionals use and evaluate such interpretable models. However, they additionally considered models with post-hoc explanations provided by \gls{shap} to compare both approaches.

In summary, when considering the focus of related work, it currently lacks a cross-model comparison to evaluate the merits and limitations of different \glspl{gam} from a neutral perspective. To close this gap, we contribute new insights with a comparative evaluation study.


\chapter{Research Method}
\label{chap:method}

To answer our research questions, we performed a series of computational experiments. In the first part, our focus was on the assessment of the predictive performance (RQ1), and subsequently, we considered the models' interpretability (RQ2). For our cross-model comparison, we examined five different \gls{gam}/\gls{gaim} approaches, for which publicly accessible implementations are available. This includes (i)~\textbf{Splines} \citep{hastie_generalized_1986}, (ii)~\textbf{EBM} \citep{nori_interpretml_2021}, (iii)~\textbf{NAM} \citep{agarwal2021neural}, (iv)~\textbf{GAMI-Net} \citep{yang_gami-net_2021}, and (v)~\textbf{ExNN} \citep{yang_enhancing_2021}. The implementations are provided by the respective authors of the proposed models, except for splines, for which we used the Python package pyGAM. Additionally, several benchmark models were included for a broader comparison, including \textbf{LR} and \textbf{\gls{dt}} as common representatives of interpretable models, and \textbf{\gls{rf}}, \textbf{\gls{gbm}}, \textbf{\gls{xgb}} and \textbf{MLP} as widely known black-box models \citep{amancio_systematic_2014, roy_performance_2019}. For their implementation, we used the corresponding Python packages from the scikit-learn library, except for \gls{xgb} where we used the XGBoost library. To provide a fair comparison, we integrated all implementations into a shared environment to run the experiments under the same conditions (i.e., workstation with single GPU NVIDIA Quadro RTX A 5000, 8 CPU cores, 24 GB VRAM and 64 GB RAM, Python 3.6.8, Tensorflow 2.3.0). Considering the choice of hyperparameters, we pursued the approach of leaving the models in their default configurations or used the settings recommended by the authors. In addition, we set the number of interactions of the EBM to 10. Further details on the implementation and hyperparameter settings are given in the appendix in Table~\ref{tab:models_overview}. 

To ensure a comprehensive evaluation, we assessed the predictive performance of all models on a wide range of benchmark scenarios. For this purpose, we looked into related evaluation studies \citep[e.g.,][]{roy_performance_2019, yang_gami-net_2021, } to identify commonly used repositories that offer publicly available dataset collections, such as Kaggle (https://www.kaggle.com/) and the UCI machine learning repository (http://archive.ics.uci.edu/ml/). We then selected appropriate datasets in such a way that their inherent properties cover a broad and representative range of real-world applications and that the prediction tasks are associated with organizational and/or societal challenges (as opposed to biological, physical or other phenomena). Furthermore, we limited our analysis to medium-sized datasets to keep the computational effort manageable. As a result, we chose twelve datasets with corresponding prediction tasks that are summarized in Table \ref{tab:datasets_overview}. They cover seven (binary) classification (CLS) and five regression (REG) tasks\footnote{The imbalance between the two prediction tasks reflects the fact that the public repositories mentioned above offer far more datasets for classification tasks than for regression tasks.}. The number of observations ranges from 205 to 103,904, and the number of predictors varies between 8 and 99 with a mixed combination of numerical and categorical features. Thus, we consider a variety of settings for assessing the models. When loading the datasets, we removed IDs and variables that cause obvious data leakage, cleaned missing values, removed categorical features with more than 25 distinct values to reduce computational complexity as we one-hot encode these, and converted the targets into a common format. Apart from that, we kept the datasets in their default structure without extensive pre-processing.

\begin{table}[h]
\resizebox{\textwidth}{!}{%
\begin{tabular}{l|l|r|c|l|l}
\multicolumn{1}{c|}{\multirow{2}{*}{\textbf{Type}}} &
  \multicolumn{1}{c|}{\multirow{2}{*}{\textbf{Dataset}}} &
  \multicolumn{1}{c|}{\multirow{2}{*}{\textbf{Observations}}} &
  \textbf{Features} &
  \multicolumn{1}{c|}{\multirow{2}{*}{\textbf{Prediction target}}} &
  \multicolumn{1}{c}{\multirow{2}{*}{\textbf{Repository}}} \\
\multicolumn{1}{c|}{} &
  \multicolumn{1}{c|}{} &
  \multicolumn{1}{c|}{} &
  \textbf{num/cat} &
  \multicolumn{1}{c|}{} &
  \multicolumn{1}{c}{} \\ \hline
\multirow{7}{*}{CLS} & Water potability                     & 3,276   & 9/0   & Will the water be safe for consumption?       & \citet{kaggle_water_nodate} \\
                     & Stroke                                & 5,110   & 3/7   & Will a patient suffer from a stroke?  & \citet{kaggle_stroke_nodate} \\
                     & Telco churn \citep{IBM}                & 7,043   & 3/16  & Will a customer leave the company?    & \citet{kaggle_telco_nodate} \\
                     & FICO credit score                     & 10,459  & 21/2  & Will a client repay within 2 years?  & \citet{fico_explainable_nodate}   \\
                     & Adult \citep{Kohavi.1996}                   & 32,561  & 7/8   & Will the income exceed \$50.000/year?    & \citet{uci_adult}    \\
                     & Bank marketing \citep{moro2014}     & 45,211  & 5/11  & Will a client subscribe to a deposit? & \citet{uci_bank}    \\
                     & Airline satisfaction                  & 103,904 & 18/4  & Will a passenger be satisfied?        & \citet{kaggle_airline_nodate} \\ \hline
\multirow{5}{*}{REG} & Car price \citep{kibler_instance-based_1989}        & 205     & 13/11 & What is the price of a car?          & \citet{uci_cars}    \\
                     & Student grade \citep{cortez_using_2008}  & 649     & 13/17 & What is a student’s final grade?     & \citet{uci_student}    \\
                     & Crimes \citep{redmond_data-driven_2002}       & 1,994   & 99/0  & How many violent crimes will happen?      & \citet{uci_crime}    \\
                     & Bike rental \citep{fanaee-t_event_2014}   & 17,379  & 6/6   & How many bikes will be rented/hour?  & \citet{uci_bike}    \\
                     & California housing \citep{pace_sparse_1997} & 20,640  & 8/0   & What is the value of a house?  & \citet{sp_sp_nodate}   \\ \hline
\end{tabular}}
\caption{Overview of used benchmark datasets for classification (CLS) and regression (REG) tasks.}
\label{tab:datasets_overview}
\end{table}

To assess the prediction qualities, we used a 5-fold cross validation where we measured the out-of-sample performance on each test fold and subsequently calculated the mean and standard deviation across all values. For classification, we measured accuracy, precision, recall, and F1-score; and for regression, we calculated \gls{rmse} and \gls{mae} as commonly applied prediction metrics. Additionally, we also measured the training times in seconds. All our experiments with the corresponding implementations and evaluation results can be found in the following GitHub repository: \url{https://github.com/fau-is/gam_comparison}.


To assess the \glspl{gam}' interpretability, we examined and compared the visual outputs of the different models. More specifically, we followed the notion that an interpretable model must be able to provide transparency at the level of the entire model (simulatability) and at the level of individual components (decomposability) to enable an understanding of how a model works \citep{lipton_mythos_2018}. To this end, we looked into the \glspl{gam}' entire output as well as specific shape functions and compared both between different models.

\chapter{Results}
\label{chap:results}

\section{Evaluation of Predictive Performance and Training Times}
\label{chap:eval_pred}

In this section, we present the results of our computational experiments. Table \ref{tab:eval_pred_cls} and Table \ref{tab:eval_pred_reg} outline the prediction results for the classification and regression tasks, respectively. We report the F1-score and RMSE as our main metrics for comparison. The best overall performance for each dataset is highlighted in bold, whereas the best result among the interpretable models is marked with an underscore. Further results on the remaining metrics can be found in our online repository.

\begin{table}[h!]
\centering
\resizebox{\textwidth}{!}{%
\begin{tabular}{l|ccccccc|cccc}
 &
  \multicolumn{7}{c|}{\textbf{Interpretable Models}} &
  \multicolumn{4}{c}{\textbf{Black-box Models}} \\
\textbf{Dataset} &
  \textbf{Splines} &
  \textbf{EBM} &
  \textbf{NAM} &
  \textbf{GAMI-Net} &
  \multicolumn{1}{c|}{\textbf{ExNN}} &
  \textbf{LR} &
  \textbf{DT} &
  \textbf{RF} &
  \textbf{GBM} &
  \textbf{XGB} &
  \textbf{MLP} \\ \hline
Water &
  .557±.020 &
  .631±.017 &
  .469±.009 &
  {\ul .634±.014} &
  \multicolumn{1}{c|}{.632±.029} &
  .464±.003 &
  .609±.007 &
  .547±.024 &
  .633±.025 &
  .642±.010 &
  \textbf{.646±.017} \\
Stroke &
  {\ul .928±.001} &
  .927±.001 &
  .928±.001 &
  {\ul .928±.001} &
  \multicolumn{1}{c|}{.927±.004} &
  {\ul .928±.001} &
  .917±.001 &
  .928±.001 &
  .928±.003 &
  \textbf{.930±.002} &
  .928±.001 \\
Telco &
  {\ul \textbf{.800±.005}} &
  .797±.006 &
  .723±.015 &
  .799±.014 &
  \multicolumn{1}{c|}{.787±.012} &
  .799±.010 &
  .747±.008 &
  .780±.017 &
  .790±.009 &
  .782±.006 &
  .790±.006 \\
FICO &
  .725±.012 &
  .725±.009 &
  .619±.065 &
  {\ul \textbf{.728±.009}} &
  \multicolumn{1}{c|}{.708±.008} &
  .718±.010 &
  .667±.015 &
  .716±.012 &
  .722±.009 &
  .710±.010 &
  .716±.011 \\
Adult &
  .854±.001 &
  {\ul .866±.001} &
  .727±.024 &
  .856±.002 &
  \multicolumn{1}{c|}{.851±.005} &
  .846±.002 &
  .846±.004 &
  .826±.004 &
  .867±.001 &
  \textbf{.868±.002} &
  .850±.003 \\
Bank &
  .893±.004 &
  .895±.002 &
  .833±.010 &
  .893±.003 &
  \multicolumn{1}{c|}{{\ul \textbf{.899±.003}}} &
  .888±.004 &
  .892±.001 &
  .859±.002 &
  \textbf{.899±.003} &
  \textbf{.899±.003} &
  .898±.004 \\
Airline &
  .935±.002 &
  .945±.002 &
  .773±.029 &
  .934±.002 &
  \multicolumn{1}{c|}{{\ul .951±.002}} &
  .875±.002 &
  .950±.002 &
  .921±.002 &
  .958±.002 &
  \textbf{.963±.002} &
  .958±.002 \\ \hline
\end{tabular}
}
\caption{Predictive performance for classification tasks measured by F1-score.}
\label{tab:eval_pred_cls}
\end{table}

\begin{table}[h]
\resizebox{\textwidth}{!}{%
\begin{tabular}{l|ccccccc|cccc}
          & \multicolumn{7}{c|}{\textbf{Interpretable Models}}                                                            & \multicolumn{4}{c}{\textbf{Black-box Models}} \\
\textbf{Dataset} &
  \textbf{Splines} &
  \textbf{EBM} &
  \textbf{NAM} &
  \textbf{GAMI-Net} &
  \multicolumn{1}{c|}{\textbf{ExNN}} &
  \textbf{LR} &
  \textbf{DT} &
  \textbf{RF} &
  \textbf{GBM} &
  \textbf{XGB}&
  \textbf{MLP} \\ \hline
Car   & .132±.022 & {\ul \textbf{.083±.0310}} & 1.112±.223 & 1.001±.100 & \multicolumn{1}{c|}{.129±.024} & .090±.033 & .148±.061 & .093±.028    & .098±.036  & .097±.038 & .106±.035    \\
Student      & .732±.129 & {\ul .730±.139} & 1.239±.161 & 1.000±.171 & \multicolumn{1}{c|}{1.390±.226} & .731±.140 & 1.239±.190 & \textbf{.712±.096}    & .757±.121 & .800±.092  & .781±.147    \\
Crimes       & .503±.021 & {\ul \textbf{.311±.026}} & 1.152±.047 & .388±.016 & \multicolumn{1}{c|}{.511±.059} & .312±.026 & .620±.056 & .319±.024    & .320±.018 & .343±.030  & .351±.038    \\
Bike      & .182±.006 & {\ul .060±.001} & 1.214±.118 & .145±.002 & \multicolumn{1}{c|}{.114±.007} & .499±.013 & .080±.003 & .202±.003    & .045±.001 & \textbf{.041±.001}  & .079±.007    \\
Housing      & .242±.008 & {\ul .181±.007} & 1.131±.153 & .266±.001 & \multicolumn{1}{c|}{.223±.008} & .352±.009 & .261±.005 & .357±.014    & .171±.006 & \textbf{.163±.007}  & .214±.007    \\ \hline
\end{tabular}}
\caption{Predictive performance for regression tasks measured by RMSE.}
\label{tab:eval_pred_reg}
\end{table}

The results demonstrate that the best prediction values for the individual datasets are highly scattered across the variety of \gls{ml} models. Taken together, the black-box models achieved the best results in $8$ out of $12$ datasets. However, with some overlap, the interpretable \glspl{gam} also achieved outperforming results in $5$ out of $12$ datasets. Additionally, it is notable that the difference between the best-performing models from both groups is marginally small. For the regression tasks, the largest difference in RMSE is $0.019$ (bike: EBM $0.06$ vs. XGB $0.041$), whereas for classification, the largest difference measured by F1-score is only $0.012$ (water: GAMI-Net $0.634$ vs. MLP $0.646$ | airline: ExNN $0.951$ vs. XGB $0.963$). These results clearly outline that there is no strict trade-off between model accuracy and interpretability.

Furthermore, EBM turns out to be the best performing \gls{gam} among all interpretable approaches showing the highest prediction qualities in $6$ out of $12$ datasets. For the regression tasks, it even outperformed all other white-box models and could achieve the best results on $2$ out $5$ datasets. For the classification tasks, GAMI-Net showed the best performance among all interpretable models, followed by ExNN and Splines. Thus, there is no single approach that clearly dominates all other interpretable models.

Considering the results of the black-box models, it is notable that XGB offers the strongest prediction qualities for a wide range of dataset properties on both types of prediction tasks. Thus, it shows the best results in 6 out 12 datasets, which justifies why the model is often the first choice among developers in \gls{ml} competitions which purely aim at achieving high prediction qualities. By contrast, it is intriguing to reveal that LR and DT, as traditional interpretable models, do not necessarily lag behind by large margins. The LR model, for example, achieves the second-highest performance in $3$ out of $12$ datasets (i.e., stroke, telco, car), affirming the observation by \citet{rudin2019stop} that the results of simple models may not significantly differ from those of black-box models when dealing with structured problems. The worst performance across the majority of datasets was achieved by the NAM model due to strong overfitting. We assume that the proposed default hyperparameters are not well-suited across various datasets. Thus, it requires further investigations in subsequent evaluation studies with a particular focus on model tuning.

Apart from the prediction qualities, we also looked into training times to assess the model usability. Table~\ref{tab:eval_traintime} provides an overview of the results measured in average seconds per fold. As expected, the simple models, i.e., LR and DT, achieve the lowest training times with values primarily $<0.1$ seconds due to their basic model structures. These models are followed by the more complex black-box models with average training times ranging from around $0.1$ up to $12$ seconds. 
Furthermore, the results of the \glspl{gam} reveal that all three neural-based approaches require much more training time than EBM and Splines, whereas the latter still show short training times up to $31$ and $39$ seconds, respectively. Among the neural-based approaches, NAM requires the most time to train, followed by GAMI-Net and ExNN. This order holds for almost all datasets across both tasks.

\begin{table}[h]
\resizebox{\textwidth}{!}{%
\begin{tabular}{l|rrrrrrr|rrrr}
          & \multicolumn{7}{c}{\textbf{Interpretable Models}}                                & \multicolumn{4}{|c}{\textbf{Black-box Models}} \\
\textbf{Dataset} &
   \multicolumn{1}{c}{\textbf{Splines}} &
 \multicolumn{1}{c} {\textbf{EBM}} &
 \multicolumn{1}{c} {\textbf{NAM}} &
  \multicolumn{1}{c}{\textbf{GAMI-Net}} &
  \multicolumn{1}{c}{\textbf{ExNN}} &
 \multicolumn{1}{|c} {\textbf{LR}} &
  \multicolumn{1}{c} {\textbf{DT}} & 
  \multicolumn{1}{|c} {\textbf{RF}} &
  \multicolumn{1}{c}{\textbf{GBM}} &
 \multicolumn{1}{c}{\textbf{XGB}} &
 \multicolumn{1}{c} {\textbf{MLP}} \\ \hline
Water      & 0.150 & 1.510 & 112.940 & 32.510 & \multicolumn{1}{r|}{11.740} & 0.020 & 0.010 & 0.220 & 0.660  & 0.110  & 0.260        \\
Stroke   & 1.230 & 0.690 & 129.110 & 46.700 & \multicolumn{1}{c|}{23.360} & 0.010 & 0.010 & 0.110 & 0.440  & 0.080      & 0.430        \\
Telco     & 2.090 & 1.510 & 270.130 & 57.510 & \multicolumn{1}{r|}{16.720} & 0.020 & 0.020 & 0.150 & 0.900 &  0.120     &   0.660  \\
FICO    & 2.370 & 1.660 & 403.550 & 95.370 & \multicolumn{1}{r|}{14.520} & 0.030 & 0.040 & 0.270 & 1.740  & 0.120      &  0.950  \\
Adult    & 38.540 & 21.640 & 1,567.940 & 519.720 & \multicolumn{1}{r|}{44.040} & 0.100 & 0.070 & 0.450        & 3.480  &   0.460   &   3.340      \\
Bank   & 15.450 & 9.490 & 1,802.360 & 777.130 & \multicolumn{1}{r|}{63.400} & 0.120 & 0.090 & 0.610 & 3.970   &  0.470   & 4.330 \\
Airline  & 17.750 & 31.020 & 3,020.080 & 627.220 & \multicolumn{1}{r|}{225.090} & 0.100 & 0.210 & 2.020  & 12.070  &  0.720    & 9.470       \\ \hline
Car   & 0.680 & 1.830 & 82.760 & 6.990 & \multicolumn{1}{r|}{8.060} & 0.001 & 0.001 & 0.080        & 0.050  &   0.040   &  0.020     \\
Student     & 0.410 & 0.440 & 70.560 & 16.010 & \multicolumn{1}{r|}{7.910}  & 0.001 & 0.001 & 0.100       & 0.090  &   0.050   &  0.060  \\
Crimes & 3.040 & 1.570 & 367.670 & 92.950 & \multicolumn{1}{r|}{9.240} & 0.001 & 0.050 & 1.320    & 1.910 &   0.120    & 0.200        \\
Bike & 0.840 & 3.820 & 557.280 & 299.090 & \multicolumn{1}{r|}{42.540} & 0.001 & 0.030 &  0.850 &  1.270  & 0.150 &   1.420 \\
Housing      & 0.180 & 6.970 & 720.800 & 418.850 & \multicolumn{1}{r|}{46.380} & 0.001 & 0.060 &   1.880  & 2.830  &  0.300    &    1.420 \\
\hline
\end{tabular}
}
\caption{Evaluation results of training time measured in average seconds per fold.}
\label{tab:eval_traintime}
\end{table}

\section{Evaluation of Model Interpretability}
\label{chap:eval_interpret}

\begin{figure}[!]
\centering
    \includegraphics[width=1\textwidth]{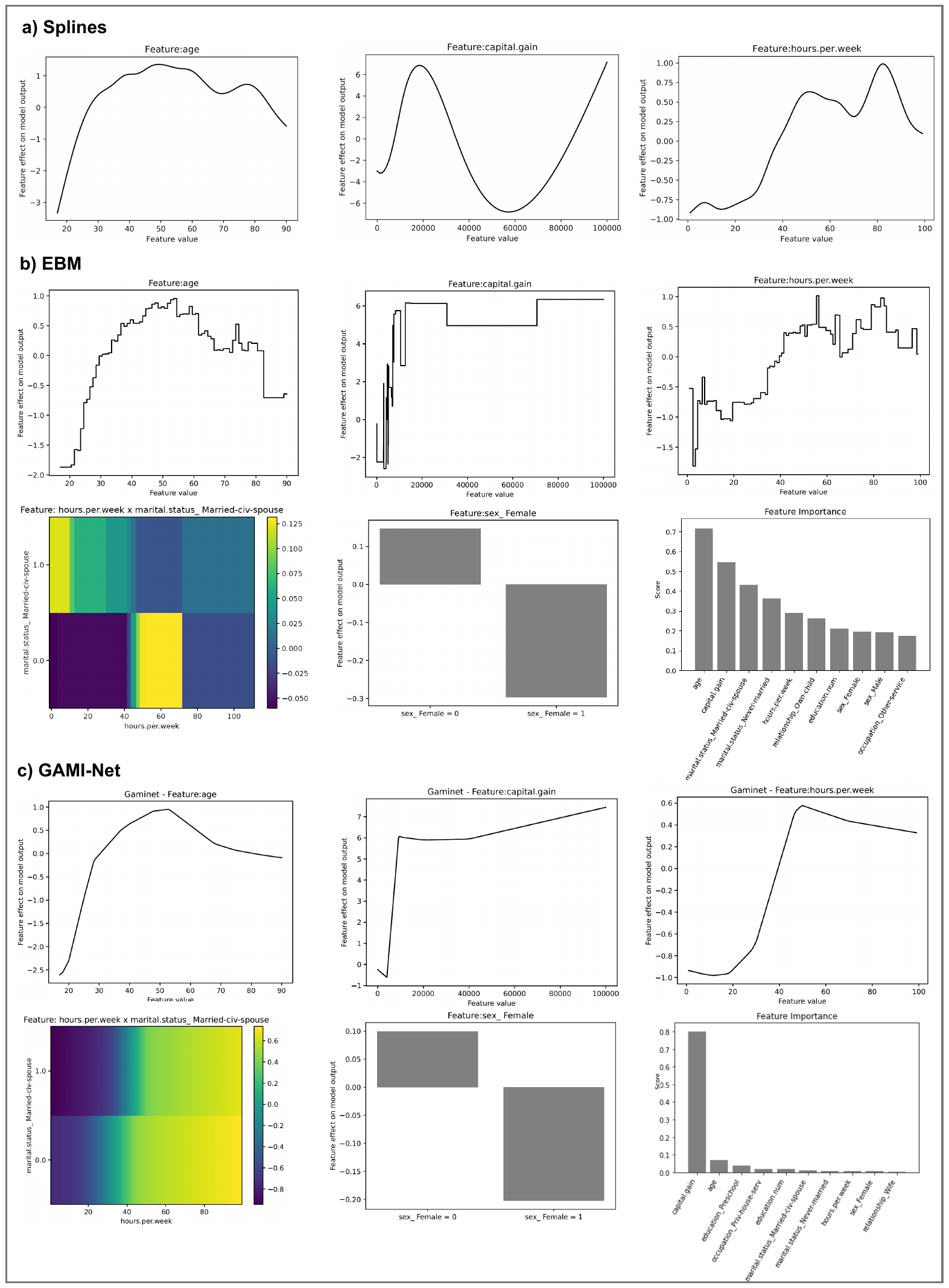}
    \caption{Visual model outputs for Splines (top), EBM (middle) and GAMI-Net (bottom).}
    \label{fig:plot1_shapefunctions_interactions_importance}
\end{figure}

\begin{figure}[!h]
\centering
    \includegraphics[width=0.9\textwidth]{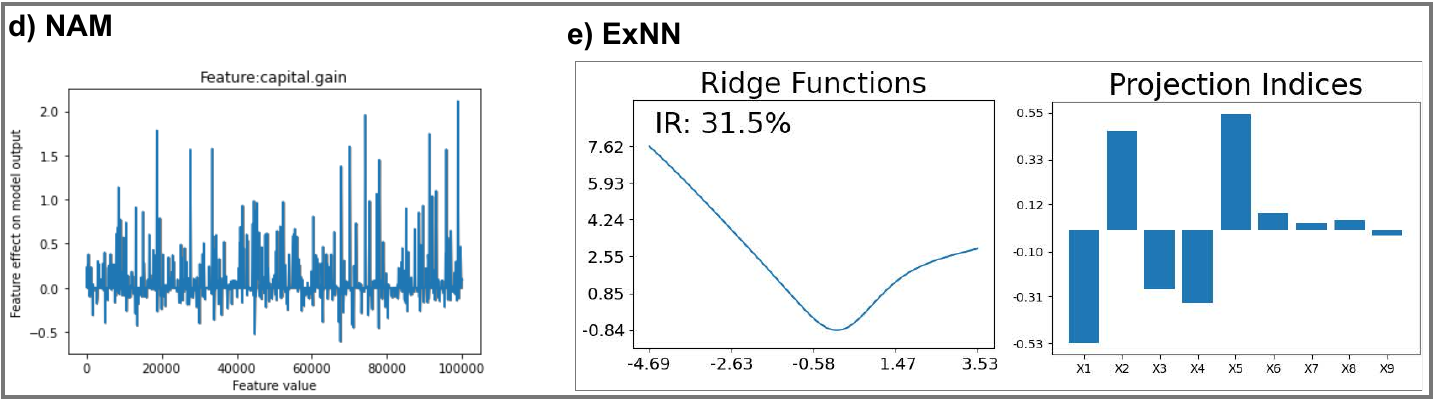}
    \caption{Visual model outputs for NAM (left) and ExNN (right).}
    \label{fig:plot2_exnn_and_nam}
\end{figure}

The qualitative assessment and discussion of the model interpretability is based on the example of the \textit{adult} dataset \citep{Kohavi.1996}. It contains about $30,000$ observations and $15$ socio-demographic and job-related features from the 1994 Census database. The prediction task is to determine whether a person makes over \$$50,000$ per year. After training all models on the entire dataset, we generated feature plots to visualize the resulting shape functions. Figure \ref{fig:plot1_shapefunctions_interactions_importance} summarizes exemplary plots for (a) Splines, (b) EBM, and (c)~GAMI-Net, whereas Figure~\ref{fig:plot2_exnn_and_nam} shows selected excerpts for (d) NAM and (e) ExNN, respectively.

Focusing on the first group of models (a-c), it can be seen that all models basically provide the same output representation, i.e., it is displayed which impact a single feature or a pair-wise feature interaction has on the target variable (i.e., \textit{income > \$50.000/year}). The impact of numerical features is shown by curves with different shapes, whereas the impact of categorical features (one-hot encoded) is shown by bar charts for the feature values $0$ vs. $1$. The x-axes represent the feature values and the y-axes represent the impact on the target variable (i.e., positive impact for $y>0$, negative impact for $y<0$). For interaction terms, both x-axis and y-axis represent the feature values, whereas the impact on the target is highlighted with a corresponding color scheme, resulting in a two-dimensional heatmap. In addition, EBM and GAMI-Net provide a calculation of global feature importance, which is displayed by bar charts with a decreasing order of relevant features. Due to space restrictions, only the ten most relevant features are shown.

To compare the same output and highlight differences between shape functions among all models, we selected a subset of important numerical features, i.e., \textit{age}, \textit{capital gain} and \textit{hours per week}, and one categorical feature, i.e., \textit{sex = female}, which is only displayed for EBM and GAMI-Net. Furthermore, since EBM and GAMI-Net can detect and capture relevant interaction terms automatically, we added an exemplary interaction plot (i.e., \textit{hours per week × marital status = married-civ-spouse}). 

From the detailed plots in Figure \ref{fig:plot1_shapefunctions_interactions_importance}, we can see that GAMs learn different patterns and relations from the training data depending on the underlying model structure. In the case of Splines (a), for example, it is observable that the feature \textit{age} has a positive effect on the target variable from a value of about $30$ years and that the effect continues to increase until about $50$ years. After this point, the impact slightly decreases again until $70$ years, where we observe another turn towards a higher effect until $80$ years. The other two models capture similar effects for that feature. However, since EBM (b) relies on an ensemble of decision-tree learners, the shape function is piece-wise constant, resulting in a step curve with sharp jumps at discrete values. On this basis, more fine-granular patterns are captured. Thus, we see more peaks with smaller ranges at certain feature values. The shape plot of GAMI-Net (c), on the other hand, shows a smooth behavior which is similar to that of Splines. Overall, such learned representations can be easily understood for interpretation purposes as they graphically reveal non-linear patterns that cannot be represented by simple LR models. Furthermore, in contrast to post-hoc-analytical methods, the plots directly display the actual decision logic of the final models without any approximations.

When comparing the other feature plots, it is also notable that the different models do not always represent the exact same patterns. Some illustrative examples are the deviating shapes for the feature \textit{capital gain}. The difference can be explained by the fact that each model uses distinct model structures (i.e., splines, vs. trees vs. neural networks) to map the feature space to the target space. This can also be seen by the different feature importance plots, where EBM due to missing sparsity constraints considers several features as almost equally important, whereas in GAMI-Net, the prediction is mostly driven by a single feature which possibly absorbs partial effects of the remaining variables.

The other two models, NAM and ExNN, were excluded from our previous considerations because they have shown a fundamentally different behavior. Due to strong overfitting effects of NAM, which we could not resolve without intensive hyperparameter tuning, the model produces extremely jagged shape functions as shown in Figure \ref{fig:plot2_exnn_and_nam} on the left side. As such, we could not generate comprehensible feature plots, which will be part of subsequent studies when tuning all models for specific data properties. The ExNN, on the other hand, is a special case as it is based on the structure of an additive index model. For the interpretation of the model, this means that not only a single feature is covered by a corresponding shape function, but that an entire set of features can possibly provide partial contributions to this shape. Figure~\ref{fig:plot2_exnn_and_nam} on the right side shows an example of such an output, which is hardly interpretable in a meaningful way if many features are involved.

\section{Summary of Cross-Model Comparison}

After examining and comparing the \glspl{gam} from different perspectives, we derive four design principles by reflecting on the merits and limitations of the individual approaches. Table~\ref{tab:models-comparison} provides an overview of the compared characteristics upon which the design principles are derived. These can serve as a summary to guide upcoming research in the field of \gls{gam} extensions and interpretable \gls{ml}. 

\begin{table}[htbp]
\centering
\resizebox{\textwidth}{!}{%
\begin{tabular}{c|c|c|c|c}
\textbf{Model} &
  \textbf{Feature Shapes} &
  \textbf{Model Compactness} &
  \textbf{Controllability} &
  \textbf{Training Effort} \\ \hline
\textit{Splines} &
  Smooth behavior &
  Low complexity &
  \begin{tabular}[c]{@{}c@{}}Controllable sparsity, \\ controllable smoothing\end{tabular} &
  \begin{tabular}[c]{@{}c@{}}Simple configuration, \\ medium training times\end{tabular} \\ \hline
\textit{EBM} &
  \begin{tabular}[c]{@{}c@{}}Piece-wise constant \\ with sharp jumps\end{tabular} &
  \begin{tabular}[c]{@{}c@{}}Increased complexity \\ due to missing sparsity\end{tabular} &
  \begin{tabular}[c]{@{}c@{}}Controllable number \\ of interactions\end{tabular} &
  \begin{tabular}[c]{@{}c@{}}Simple configuration, \\ fast training times\end{tabular} \\ \hline
\textit{NAM} &
  Jagged behavior &
  \begin{tabular}[c]{@{}c@{}}Low complexity\end{tabular} &
  \begin{tabular}[c]{@{}c@{}}Controllable regularization\\ for limiting jagged behavior\end{tabular} &
  \begin{tabular}[c]{@{}c@{}}Extensive HP tuning, \\ slow training times\end{tabular} \\ \hline
\textit{GAMI-Net} &
  Smooth behavior &
  Low complexity &
  \begin{tabular}[c]{@{}c@{}}Controllable sparsity, \\ controllable smoothness, \\ controllable interactions\end{tabular} &
  \begin{tabular}[c]{@{}c@{}}Moderate HP tuning, \\ moderate training times\end{tabular} \\ \hline
\textit{ExNN} &
  \begin{tabular}[c]{@{}c@{}}Smooth behavior \\ of feature projections \end{tabular} &
  \begin{tabular}[c]{@{}c@{}}High complexity due \\ to feature projections\end{tabular} &
  \begin{tabular}[c]{@{}c@{}}Controllable sparsity,\\ controllable smoothness\end{tabular} &
  \begin{tabular}[c]{@{}c@{}}Moderate HP tuning, \\ moderate training times\end{tabular} \\ \hline
\end{tabular}
}
\caption{Comparison of examined \gls{gam} characteristics.}
\label{tab:models-comparison}
\end{table}

\textit{\textbf{Principle 1:} Provide constraints to derive shape functions that can capture complex patterns but remain comprehensible}. As shown before, simple Splines have the advantage of being simply comprehensible due to their strict smoothness, but this can also impair predictive performance as they cannot capture more complex patterns. NAM and EBM, on the other hand, are able to extract and depict non-smooth effects, such as sharp jumps due to abrupt changes or discrete thresholds given in real-world settings. Nevertheless, this behavior also bears the risk of producing unexpected jumps that are difficult to justify, which may become even worse in the presence of noisy samples and outliers. GAMI-Net offers a compromise by providing smooth shape outputs while still guaranteeing strong prediction qualities based on neural feature subnets and the identification of relevant pair-wise interactions. To take account for different patterns and shapes, a future innovation could be, that a model is able to capture different shapes, depending on whether a user prefers strictly smooth curves or jagged behavior with sharp jumps.

\textit{\textbf{Principle 2:} Provide constraints to keep the model's complexity manageable.} The ExNN is an example of a model where the internal decision logic is transparent but difficult to comprehend --- especially for real-world problems where many features are involved. Thus, although the additional projection layer may boost predictive performance, it is hard to explain the practical meaning of partial feature combinations that are spread over multiple shape functions. The other models, on the other hand, only depict a single feature at a time --- or at most a pair-wise interaction --- which remains simply comprehensible. Furthermore, to keep model complexity manageable, some models such as GAMI-Net and Splines provide sparsity constraints to receive a more compact representation that concentrates on fewer features. Since EBM does not provide such constraints, the overall model complexity cannot be controlled.

\textit{\textbf{Principle 3:} Provide constraints to control the model's overall structure and its components.} All five approaches provide some sort of controllability to modify the model's structure, such as the number of interactions (e.g., EBM), smoothness (e.g., ExNN), and/or sparsity (e.g., GAMI-Net). Such control appears to be crucial when model developers and domain experts want to integrate their knowledge into the model structure, depending on the situation. For example, by controlling the sparsity, a developer can counteract the problem of multicollinearity in case of highly correlated but irrelevant features. By contrast, a non-sparse model might help to identify biases within the training data which in case of a strong default regularization could be compiled into other correlated features and remain unrecognized \citep{chang_how_2021}. Therefore, it is helpful to provide different mechanisms for model modification so that users can create models according to their specific needs and intentions.

\textit{\textbf{Principle 4:} Provide mechanisms to allow for simple usability with minimal training effort.} While EBM and Splines are comparatively fast and can be simply applied without much configuration effort, the neural-based GAMs suffer from the limitation that they are computationally intensive and require careful parameter tuning. Especially for NAM, it appears highly critical to choose a suitable set of hyperparameters (HP) that regulate the overfitting effect. Consequently, such approaches lack user-friendly applicability so that a model can be quickly applied without much training effort. This seems essential for interpretable models since users and developers may go through several iterations of model adjustments after spotting the performance and the visual output of individual shape functions.

\chapter{Discussion and Outlook}
\label{chap:discussion}

\gls{ml} models are highly beneficial for capturing hidden patterns and crucial nuances in large datasets. It can therefore be expected that the importance and prevalence of \gls{ml} models will continue to increase in the future to assist or complement human decision-making and prediction tasks. Thus, the central question will not be \textit{whether} \gls{ml} models should be employed, but instead \textit{which} type of \gls{ml} models. In several domains, such as healthcare or financial risk assessment, simple regression models are still the most widely applied models as transparency is mandated due to the severe consequences of incorrect model predictions or because regulatory requirements enforce it \citep[e.g.,][]{valaskova_financial_2018, Bertoncelli.2020, Shipe2019}. At the same time, many research communities currently emphasize the development and application of XAI approaches to make complex black-box models more transparent and comprehensible.
A similar development can be observed in our \gls{is} discipline, with an increasing focus on post-hoc-analytical methods such as LIME or SHAP \citep[e.g.,][]{wanner_white_2020, wanner_how_2020, wastensteiner_explainable_2021, mehdiyev_local_2021, jussupow_is_2021,zhang_explainable_2020, schemmer_intelligent_2021, stierle_bringing_2021}. However, the concerns being raised about post-hoc-analytical methods should also be taken seriously in our field \citep{rudin2019stop}. The option of choosing a complex black-box model, which subsequently requires a posteriori explanations, should only be considered if there is no better alternative to it. To this end, \citet{rudin2019stop} emphasizes the need in critical applications to first prove that high predictive performance can only be provided by black-box models, which in turn cannot be achieved with interpretable alternatives.

The results of our study directly contribute to this central debate. Thus, we were able to show that intrinsically interpretable \gls{ml} models can indeed achieve competitive prediction qualities, providing further evidence that there is no strict trade-off between model interpretability and model accuracy. In fact, the performance of the examined \glspl{gam} was much closer to the prediction results of the black-box models than to those of the traditional white-box models, if not even outperforming them in $5$ out of $12$ prediction tasks. Consequently, we argue that advanced \gls{gam} models such as EBM or GAMI-Net should be firmly established as first-choice models in predictive modeling projects as we see large potential of these models to change the game when developing and applying \gls{ml} approaches in research and industry alike.

Against this background, we also see an important role of these models in our socio-technical discipline of \gls{is} research. Due to their flexibility to capture complex patterns and their simplicity to produce easily understandable outputs, they provide a technically equivalent, but ethically more acceptable alternative to black-box models. 
In the past, we saw many examples where bias issues and fairness problems have been reported in \gls{ml} applications. Prominent examples stem from recruiting applications where women have been discriminated or in pretrial detention and release decisions with skewed predictions against African Americans \citep{janiesch_machine_2021, mehrabi_survey_2021}. While detecting biases in training data is considered a tedious task, with advanced \glspl{gam} such distortions can be better spotted by means of the shape functions within the final model in order to avoid racial, sexual and other kind of discrimination. For instance, taking our demonstration case from Section \ref{chap:eval_interpret} as an example, it can be quickly spotted (i)~which model incorporates critical features such as gender, age and marital status for the income prediction, (ii)~which feature values have a stronger tendency towards a higher income (e.g., sex = male, age between 30 and 60), and (iii)~whether the models captured interaction effects of critical features with other variables that need to be considered (e.g., marital status × hours per week).

Likewise, model users can better interact with such intrinsically interpretable \gls{ml} models, since they directly see how individual features influence a model's outcome. Thus, domain knowledge can be incorporated into the model by removing or adding certain features, adjusting specific model properties (e.g., smoothness), or adding further model constraints (e.g., upper and lower bounds in shape functions) to take better account for desired patterns. Because of all these properties, \glspl{gam} can be considered promising for critical applications in practice.
For example, in the health sector, full transparency of diagnoses is essential for medical professionals to reconcile the model's results with their own experience and build trust in the prediction.
Other examples may refer to safety-critical applications, such as natural disaster detection, crime prediction and financial fraud detection, in which it is crucial to gather a full understanding about a model's predictor variables and their impact on the target.

Nevertheless, to avoid any confusion, we also want to emphasize the point that careful conclusions have to be drawn on the basis of such interpretable models. Using the words of \citet{caruana2015intelligible}, \textit{\textquote{it is tempting to interpret them causally}}, but even though the evaluated \glspl{gam} offer transparent explanations of how predictions are derived, they are still based on correlations. As such, it cannot be said for certain why some of the effects shown in the feature plots are present. This could be due to overfitting, interactions and correlations with other (unmeasured) variables, or due to other underlying phenomena. Nevertheless, even though such ML models cannot guarantee causality by themselves, they provide some suggestive indications for further investigations that can be helpful for the identification of causal pathways. 

As with any research, our work is not free of limitations. In the current study, we restricted our analyses to medium-sized datasets to keep the computational costs manageable. This was necessary as we could see that almost all \gls{gam}-based models require high computational resources with large training times. Thus, investigating datasets with other properties (e.g., large-scale collections with several million observations and many more features) is an open issue for future work to provide more evidence for our findings.

Likewise, we explicitly focused on prediction tasks with tabular data that usually contain naturally meaningful features for interpretation purposes. In domains with higher-dimensional data such as images, text, and event logs, the results of this study may not directly apply. For this purpose, some upstream feature engineering methods are required that transform raw input data like signals, pixels or text snippets into higher-level features before feeding them into \glspl{gam} to produce interpretable shape functions.

Furthermore, we refrained from performing extensive hyperparameter tuning for each model and dataset. To this end, it can be assumed that both the traditional models as well as the \gls{gam}-based approaches can achieve even higher prediction performance. However, since we worked mostly with default settings in both groups and avoided explicit optimizations, the comparison can be considered fair. In subsequent work, we will conduct further experiments on this open issue.

A last limitation concerns the evaluation of the model interpretability. As a first step, we tried to pursue an objective evaluation of the learned shape functions of interpretable models to derive a better understanding of the various models and their properties before moving on to a socio-technical investigation in the next step. For this purpose, we compared the outputs of different \glspl{gam} qualitatively, and derived statements about their merits and limitations. Thus, the evaluation of the subjective perception by model users, in which all \glspl{gam} with their characteristic outputs are evaluated in realistic decision-making scenarios, is currently missing. To this end, it is planned to conduct field experiments with data science experts and decision-makers from different domains and evaluate the usefulness of the different models using real-life prediction~problems. 

Another avenue for future research is to investigate under which circumstances one interpretable model should be preferred over another. To this end, our initial findings on the merits and limitations of the different \gls{gam} models provide a valuable starting point for identifying more specific selection criteria and deriving recommendations as to which model might be more appropriate in a given situation.

\chapter*{Appendix}
\label{chap:appendix}

\begin{table}[h]
\resizebox{\textwidth}{!}{%
\begin{tabular}{@{}lll@{}}
\toprule
\textbf{Models} & \textbf{Python Implementations}            & \textbf{Hyperparameter Setting}          \\ \midrule
Splines         & \texttt{pyGAM}[.LogisticGAM/.LinearGAM] \hspace{0.05cm} \citep{serven_pygam_2021}                          & Default                                       \\
EBM             & \texttt{interpret}.ExplainableBoosting[Classifier/Regressor] \hspace{0.05cm} \citep{nori_interpretml_2021}                      & Interactions = 10                                       \\
NAM             & \texttt{nam} \hspace{0.05cm} \citep{google_research_nam_2021}                            & Default                                       \\
GAMI-Net        & \texttt{gaminet} \hspace{0.05cm} \citep{yang_gami-net_2021-git}                       & Default                                       \\
ExNN            & \texttt{exnn} \hspace{0.05cm} \citep{yang_enhanced_2021}                           & Default                                       \\
LR              & \texttt{sklearn.linear\_models[LogisticRegression/Ridge]}             & L2 regularization (logistic), ridge regression (linear)\\
DT              & \texttt{sklearn.tree}.DecisionTree[Classifier/Regressor]      &  Max depth = 12                                            \\
RF              & \texttt{sklearn.ensemble.RandomForest[Classifier/Regressor]}     & Max depth = 5, number of estimators = 100                            \\
GBM             & \texttt{sklearn.ensemble.GradientBoosting[Classifier/Regressor]} & Max depth = 5, number of estimators = 100                        \\
XGB             & \texttt{xgboost.XGB[Classifier/Regressor]} & max depth = 5, number of estimators = 100                        \\
MLP             & \texttt{sklearn.neural\_network.MLP[Classifier/Regressor]}       & \begin{tabular}[c]{@{}l@{}}Number of hidden layers = 1, number of neurons = 40,\\ number of epochs = 100, activation function = ReLU\end{tabular}                                              \\ \bottomrule
\end{tabular}}
\caption{Overview of applied models with implementations and configurations.}
\label{tab:models_overview}
\end{table}


\renewcommand*{\chaptitlefont}{\fontsize{14}{10}\selectfont\sffamily}
\printbibliography{}

\end{document}